\documentclass{Interspeech}

\interspeechcameraready

\title{Iterative refinement, not training objective, makes HuBERT behave differently from wav2vec 2.0}

\author[affiliation={1}]{Robin}{Huo}
\author[affiliation={1,2,3}]{Ewan}{Dunbar}

\affiliation[]{Department of Linguistics}{University of Toronto}{Canada}
\affiliation[]{Department of Computer Science}{University of Toronto}{Canada}
\affiliation[]{Department of French}{University of Toronto}{Canada}
\email{robin.huo@mail.utoronto.ca, ewan.dunbar@utoronto.ca}

\usepackage{comment}

\begin{document}

\maketitle

\begin{abstract}

    Self-supervised models for speech representation learning now see widespread use for their versatility and performance on downstream tasks, but the effect of model architecture on the linguistic information learned in their representations remains under-studied. This study investigates two such models, HuBERT and wav2vec 2.0, and minimally compares two of their architectural differences: training objective and iterative pseudo-label refinement through multiple training iterations. We find that differences in canonical correlation of hidden representations to word identity, phoneme identity, and speaker identity are explained by training iteration, not training objective. We suggest that future work investigate the reason for the effectiveness of iterative refinement in encoding linguistic information in self-supervised speech representations.

\end{abstract}

\section{Introduction}

The use of self-supervised learning (SSL) methods for learning speech representations has become commonplace in recent years, owing to its improvement of downstream performance, reduction of the amount of labelled data needed to develop task-specific models, and versatility in a variety of applications. However, while the utility of SSL representations for downstream tasks has been well shown \cite{yang_superb_2021}, and while it is understood that many forms of speech SSL learn meaningful linguistic representations \cite{martin_probing_2023, pasad_comparative_2023, pasad_what_2024}, the reasons why they succeed have remained unclear. In particular, the effect of choices in model architecture and training regime on the linguistic information and structure in their representations remains under-studied.

Of the SSL models for speech in popular use today, most are based on the architecture of HuBERT \cite{hsu_hubert_2021} or wav2vec 2.0 \cite{baevski_wav2vec_2020}. Recent work has shown that these two models show appreciable differences with respect to the linguistic information in their learned representations \cite{pasad_comparative_2023, pasad_what_2024, choi_self-supervised_2024, sanabria_analyzing_2023}.


Despite these behavioural differences, HuBERT and wav2vec 2.0 have only a few important differences in architecture. First, while both models predict pseudo-labels of masked input frames, HuBERT optimizes a masked language modelling classification objective, whereas wav2vec 2.0 optimizes a contrastive objective with negative examples. Second, HuBERT obtains the pseudo-labels for this task from k-means clustering on acoustic features or existing HuBERT representations and keeps them fixed, whereas wav2vec 2.0 jointly learns its pseudo-labels using a quantization module. Finally, HuBERT is pretrained in multiple iterations with each trained iteration providing representations to be clustered into the pseudo-label categories of the next, while wav2vec 2.0 is pretrained only once.

We isolate and investigate the effect of the training objective and iterative refinement of pseudo-labels via multiple training passes on the linguistic information encoded in the resulting pretrained model.
We find that the critical difference is the use of iterative refinement.
The behaviour of a HuBERT-like model with respect to layer-wise correlation of its representations to words and phonemes is predicted by the number of training iterations, not by training objective: as the training iteration increases, so does the level of linguistic correlation in the final layers.
Furthermore, this improvement in linguistic encoding across iterations appears to come at the cost of non-linguistic speaker information.
On the basis of these results, we suggest that future research investigate the reason for the effectiveness of iterative refinement in producing linguistically meaningful representations, and develop ways to leverage this insight for more efficient training of high-quality representation learners.

Code and checkpoints for our investigation can be found at \url{https://github.com/RobinHuo/iter-ref}.

\section{Preliminaries}

\subsection{HuBERT and wav2vec 2.0}
HuBERT \cite{hsu_hubert_2021} and wav2vec 2.0 \cite{baevski_wav2vec_2020} are popular SSL speech models which have inspired many subsequent variants and architectures \cite[for example]{chen_wavlm_2022, babu_xls-r_2022, chung_w2v-bert_2021}.
The \textsc{base} variants of HuBERT and wav2vec 2.0 consist of a 7-layer convolutional waveform encoder, followed by a 12-layer bidirectional Transformer encoder \cite{devlin_bert_2019}, and a final projection.
In pretraining, both models use a masked prediction task with a loss of the form
\[
\mathcal{L}_\text{masked} =
-\log\frac{\exp(\operatorname{sim}(x, e_c)/\tau)}{\sum_{c' \in C} \exp(\operatorname{sim}(x, e_{c'})/\tau)}
\]
where $\operatorname{sim}(\cdot,\cdot)$ is cosine similarity, $c$ is the pseudo-label of the masked frame being predicted, $x$ is the output of the model for that frame, $e_p$ is the embedding for pseudo-label $p$, $\tau$ is a temperature hyperparameter, and $C$ is the set of pseudo-labels from which the model must predict the correct choice.
The set $C$ differs between HuBERT and wav2vec 2.0:
in HuBERT this is the set of all pseudo-labels, whereas in wav2vec 2.0 it is a set consisting of the correct pseudo-label $c$ along with $K$ negative examples sampled from the pseudo-labels of other masked frames in the same utterance, where $K$ is a hyperparameter.
The loss used by wav2vec 2.0 is termed a \emph{contrastive} loss, as it is designed to make the model learn to distinguish between frames in the same utterance. The HuBERT loss is termed \emph{predictive.}

The pseudo-labels used by HuBERT and wav2vec 2.0 are computed differently.
In HuBERT, the pseudo-labels remain fixed for the duration of training, are subsequently updated, and training is then restarted with a newly initialized model (iterative refinement).
The first training iteration uses pseudo-labels from clustered acoustic features of the input while subsequent iterations use pseudo-labels from clustered hidden representations yielded by the previous training iteration.
In contrast, wav2vec 2.0 undergoes only one training pass using an online quantization module (2-codebook product quantization learned with Gumbel softmax) to jointly learn a set of pseudo-labels.
In order to ensure full use of this quantization module, wav2vec 2.0 also employs a diversity loss in addition to the masked prediction loss ($\mathcal{L} = \mathcal{L}_\text{masked} + \alpha \mathcal{L}_\text{diversity}$, where $\alpha$ is a hyperparameter), which encourages equal use of all codewords.

\subsection{Canonical correlation analysis}
We follow Pasad et al.\ \cite{pasad_layer-wise_2021, pasad_comparative_2023, pasad_what_2024} in using canonical correlation analysis (CCA) to evaluate the linguistic information contained in SSL representations.
CCA is a technique for characterizing the relationship between two random vectors in terms of correlations between linear combinations of their features.
Given $n$ pairs of vectors $(x, y)$ sampled from the random vectors $X \in \mathbb{R}^{d_x}$ and $Y \in \mathbb{R}^{d_y}$, CCA computes $\min\{d_x,d_y\}$ canonical variable pairs $U_i = u_i^\mathsf{T} X$ and $V_i = v_i^\mathsf{T} Y$ where $u_i \in \mathbb{R}^{d_x}$ and $v_i \in \mathbb{R}^{d_y}$,
such that the Pearson correlation $\operatorname{cor}\left(U_i, V_i\right)$ is maximized for each $i$ subject to the constraint that each canonical variable $U_i$ and $V_i$ is uncorrelated with both $U_j$ and $V_j$ for all $j < i$.
The quantity $\operatorname{cor}\left(U_i, V_i\right)$ is the $i$-th canonical correlation.
Projection-weighted CCA \cite{morcos_insights_2018} is a variant of CCA which returns a weighted sum of the canonical correlations such that directions accounting for higher proportion of the input receive higher weight.
Intuitively, the scalar value returned by projection-weighted CCA is an integrated measure of similarity, as measured by canonical correlations, between two multivariate data series of interest.
When comparing model representations to linguistic categories using this method, we henceforth refer to this value as the \emph{CCA score.}

\section{Influence of model architecture and training on linguistic encoding}
We investigate the effect of specific factors in self-supervised speech model architecture and training on the linguistic information encoded in the hidden representations of the resulting pretrained model, focusing on HuBERT and wav2vec 2.0.

We focus on the fact that the encoding of information about phoneme and word identity as measured by CCA shows a marked decrease in the final layers of wav2vec 2.0 that is not seen in HuBERT, with this decrease being more acute for words than phonemes \cite{pasad_comparative_2023, pasad_what_2024}. This difference in behaviour is present across both the \textsc{base} and \textsc{large} variants of these models.

It has been suggested that the choice of pretraining objective is a significant factor in the amount and nature of linguistic information encoded by the resulting model \cite{chung_similarity_2021, chung_w2v-bert_2021, choi_self-supervised_2024, pasad_what_2024}.
To determine whether the training objective or difference in pseudo-label refinement strategies between HuBERT and wav2vec 2.0 causes the difference in behaviour, we minimally modify HuBERT to use wav2vec 2.0's contrastive pretraining objective, train the resulting model for two iterations, and compare its behaviour to two iterations of standard HuBERT with respect to canonical correlation to phoneme and word identity, following the method of Pasad et al.\ \cite{pasad_comparative_2023}.

Previous work has also shown a difference in encoding of speaker information between HuBERT and contrastive models such as wav2vec 2.0 and CPC \cite{mohamed_orthogonality_2024, peng_probing_2024, polyak_speech_2021, van_niekerk_acoustic_2024}.
In order to evaluate the effect of our manipulations on non-linguistic speaker information, we extend the CCA analysis to speaker identity.

\section{Experimental setup}
\subsection{Models}
\label{models}
To implement our modified contrastive HuBERT (henceforth cHuBERT), we used the fairseq \cite{ott_fairseq_2019} implementation of HuBERT-\textsc{base} (90M parameters) \cite{hsu_hubert_2021} as a base and modified it by integrating the code for the contrastive loss from the fairseq implementation of wav2vec 2.0 \cite{baevski_wav2vec_2020}.
We then pretrained four models: first-iteration HuBERT, first-iteration cHuBERT, second-iteration HuBERT, and second-iteration cHuBERT. Following the original HuBERT paper \cite{hsu_hubert_2021}, our first-iteration models used k-means cluster pseudo-labels from MFCCs (13 coefficients + first- and second-order derivatives) with $k = 100$. The second-iteration models used k-means clusters on representations from the 6th Transformer layer of our first-iteration HuBERT with $k = 500$. The number of negative samples for the contrastive loss in the cHuBERT models was set to 100, in line with wav2vec 2.0. Note that cHuBERT is different from wav2vec 2.0 in that it does not learn pseudo-labels jointly.

All models were pretrained for 250k updates on LibriSpeech-960h, a set of public English audiobooks \cite{panayotov_librispeech_2015}. Following the original HuBERT paper, the learning rate was linearly ramped up from 0 to a peak of $5 \times 10^{-4}$ for the first 8\% of updates and linearly decayed to 0 over the remainder of training. We used 4 GPUs with 8-step gradient accumulation and the default per-device batch size of 1.4M frames to emulate the 32-GPU training setup of the original paper. All other hyperparameters were left untouched from the defaults in fairseq.

As a follow-up to confirm the effect of iteration, we also trained a third-iteration HuBERT model using k-means clusters from the 9th Transformer layer of our second-iteration HuBERT model with $k = 500$. The training setup was as above. As a second follow-up to test whether the observed differences were due to iterative refinement or cumulative training time, we trained a wav2vec 2.0-\textsc{base} model (95M parameters) for 500k updates, twice the amount of cumulative training of one (c)HuBERT iteration, using the fairseq implementation.
The other training details were as above. Finally, as a random baseline, we included a randomly initialized (untrained) HuBERT-\textsc{base} model.

\subsection{CCA analysis of linguistic and speaker information}
We followed the method and implementation of Pasad et al.\ \cite{pasad_comparative_2023} to evaluate the above models for layer-wise CCA correlation to phoneme identity and word identity on LibriSpeech dev-clean. We additionally added an analogous analysis for speaker identity, which treated each utterance as a speaker token in accordance with work showing that the per-utterance mean yields a good representation for speaker information \cite{van_niekerk_analyzing_2021, van_niekerk_acoustic_2024}.

For each model, we computed the CCA score of the hidden representations of each layer with one-hot vectors encoding phoneme, word, or speaker identity for a sample of the data. Following \cite{pasad_comparative_2023}, phoneme tokens were mean-pooled across the middle third of the hidden representation frames and word tokens were mean-pooled across frames for the whole token. Speaker tokens were mean-pooled across the whole utterance.

\begin{figure*}[ht!]
    \centering
    \includegraphics[width=1\linewidth]{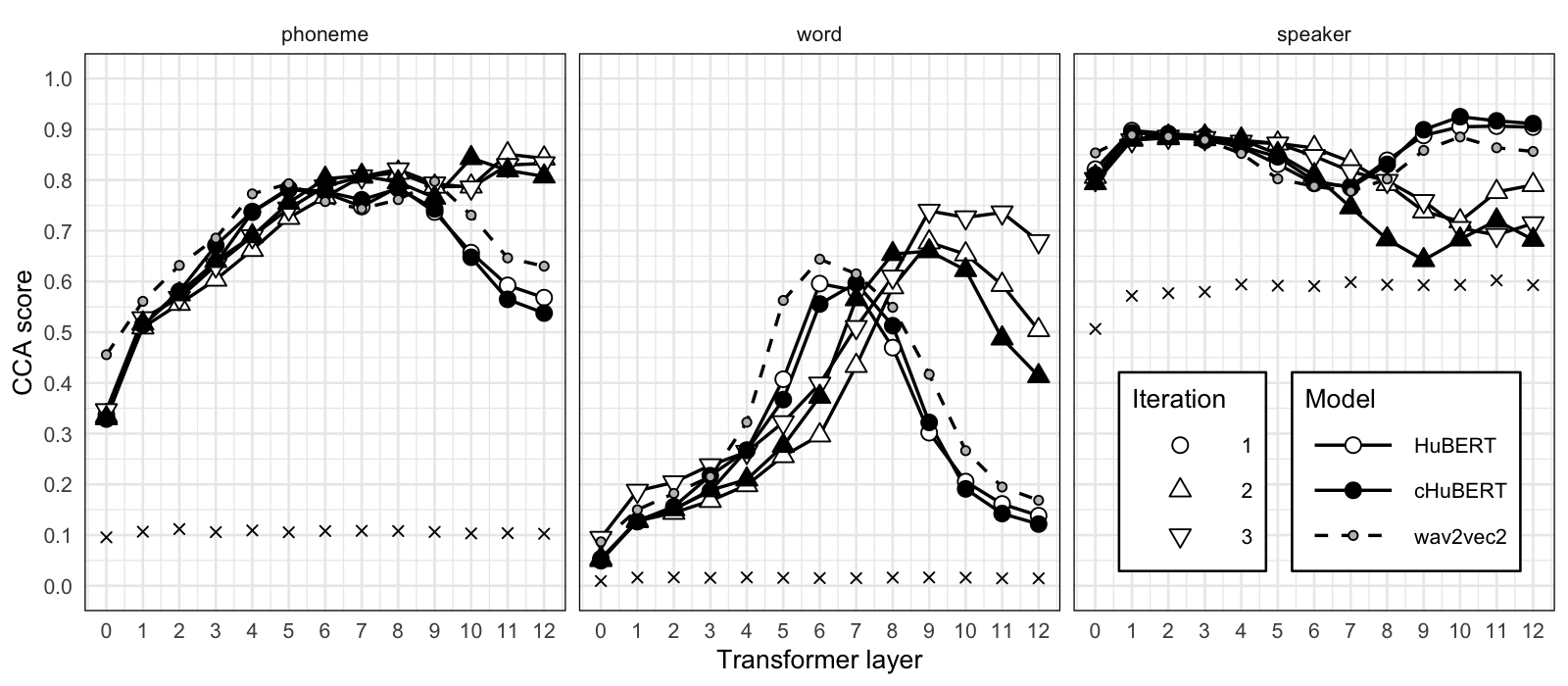}
    \caption{Layerwise CCA scores for all tested models with respect to phoneme identity, word identity, and speaker identity. Transformer layer 0 denotes the input to the first Transformer layer. The randomly initialized baseline is plotted with the symbol $\times$.}
    \label{fig:cca}
\end{figure*}

We followed the procedure of \cite{pasad_comparative_2023} for data sampling and CCA model fitting.
A maximum of 200 tokens per phoneme type was sampled across 39 phoneme types, for a maximum of 7800 tokens.
A maximum of 15 tokens per word type was sampled across 500 word types, for a maximum of 7500 tokens.
For speakers, all utterances were used.
The CCA score was calculated across three different cross-validation splits (train--dev--test) over the total sample for tuning the regularization hyperparameters ($\epsilon_x$ and $\epsilon_y$).
This sampling was conducted three times. Each reported CCA score is an average of nine projection-weighted CCA models: three cross-validation splits for each of three data samples.

\section{Results}
\subsection{Effect of training objective and iterative refinement}
Figure \ref{fig:cca} shows the CCA scores of the first two iterations of HuBERT and cHuBERT with phoneme, word, and speaker identity.
Beginning from the middle layers, large differences can be seen between iterations for all three CCA scores.

Past the 9th layer, correlation with phoneme identity falls drastically for the first-iteration but not the second-iteration models.
Correlation with word identity for the first-iteration models peaks at layers 6 and 7 and falls to finish at near-initial levels.
In contrast, the second-iteration models peak later (layer 9 vs.\ layers 6/7) and higher (around 0.66 vs.\ 0.6), and do not fall nearly as dramatically in the final layers compared to the first-iteration models.
Correlation with speaker identity displays the opposite pattern:
the first-iteration models attain their highest value at the final layers, whereas the second-iteration models display a prominent drop in correlation at the final layers.

Overall, these results show a marked increase in correlation to word and phoneme identity and a decrease in correlation to speaker identity in the final layers from the first iteration to the second iteration of training.
The choice of training objective shows no comparable effect.
Note that these results reproduce the pattern found by Pasad et al.\ \cite{pasad_comparative_2023} for phoneme and word identity in HuBERT and wav2vec 2.0, with our first-iteration models showing the pattern of their wav2vec 2.0 model.
The crucial difference is thus the use of iterative refinement in HuBERT.

\subsection{Iterative refinement vs.\ cumulative training exposure}
Alternately, one might think that the crucial factor is not the technique of iterative refinement per se but rather the total amount of training invested in the model, as the second-iteration models rely on the training of the first-iteration models.
To test this, we evaluated a wav2vec 2.0-\textsc{base} model trained for 500k updates, the same total number as one of our first- and second-iteration (c)HuBERT models combined.
The dashed line in Figure \ref{fig:cca} shows the result of the CCA evaluations on this wav2vec 2.0 model.
The wav2vec 2.0 model patterns with the first-iteration models and not the second-iteration models, confirming that the critical difference is the iterative refinement of pseudo-labels when passing between training iterations.

\subsection{Generalization of effect of training iteration}
In order to test the generalization that correlation with linguistic categories increases and correlation with non-linguistic categories decreases with training iteration, we evaluated a third-iteration HuBERT model (details in \S\ref{models}).
Figure \ref{fig:cca} shows CCA scores with phoneme, word, and speaker identity across three iterations of HuBERT (white points).
Compared to the second iteration, the third-iteration HuBERT attains a higher peak and a higher CCA score in the final layers for word correlation.
Correlation with phoneme identity shows no appreciable difference from the second iteration but does not decrease.
Correlation with speaker identity degrades in the final layers, as expected.

\section{Discussion}
The results suggest a strong role for iterative refinement of pseudo-labels and minimal influence of training objective on the phoneme and word information encoded in hidden representations of HuBERT-like models.
More rounds of iterative refinement results in more informative representations of words and phonemes in the final layers.
Furthermore, the improved representation of words and phonemes across iterations appears to come at the cost of non-linguistic information, in particular speaker identity, which shows decreased CCA scores across iterations.
These results accord with findings that performance of HuBERT on high-level linguistic tasks improves across three iterations \cite{nguyen_are_2022}.
Our results also reinforce the message that the nature and content of the pseudo-labels in pretraining is a critical factor in the use of SSL speech models \cite{nguyen_are_2022, ma_pushing_2023}.

We hypothesize that the intermediate layers learn and operationalize abstractions, focusing some kinds of information and backgrounding or erasing others according to what is useful for doing the categorization specified by the pseudo-label targets.
Iterative refinement uses clusters of these abstracted representations as pseudo-labels in subsequent models.
This begets further abstraction in favour of the information focused by those representations.
This is reflected in our results by the increase of phoneme CCA scores after the first iteration, the increase of word CCA scores across all three iterations, and the decrease of speaker CCA scores across all three iterations in the final layers.

Why do these models learn about phonemes and words but not speakers?
The specific categories which the model chooses to abstract are a function of the data, the pseudo-labels, and the task.
We may expect that speaker information would be backgrounded or discarded for at least two reasons.
First, the input pseudo-labels for the first iteration are clusters of mel-frequency cepstral coefficients (MFCCs),
which are designed to disentangle speaker information from linguistic information \cite{picone_signal_1993}.
Typical use, including in the training of HuBERT, involves retaining just the linguistically relevant (lower-order) coefficients.
Second, the typical training regime of HuBERT does not involve comparisons across speakers.
Following the original paper, we trained HuBERT-\textsc{base} on LibriSpeech, a collection of public English audiobooks.
Crucially, each example in this dataset contains only a single speaker.
This means that the sequences encountered by HuBERT during pretraining each contain information from only a single speaker, so no attention is ever computed between different speakers and HuBERT cannot directly form an abstract model of speaker variation.

On the other hand, while abstract linguistic units such as phonemes and words form an important low-bitrate representation of language, and thus it makes sense that they would be useful for prediction, the first-iteration clusters from MFCCs will not align directly with word or phoneme categories.
The first-iteration model will thus show a degradation in linguistic correlation as it approaches the last layer in order to enhance the linguistically irrelevant information captured by these targets.
This may be seen as a task effect.
Previous work has suggested that in many cases, fine-tuning for a specific task primarily and significantly changes the representational content of the final layers \cite{pasad_layer-wise_2021, yang_what_2023}.
This supports the view of the final layers as a task-specific transducer converting from abstract representations in the intermediate layers, which are relatively invariant as they model inherent and fundamental structure in the data, to surface-level features required for a concrete task.

\subsection{Implications for training and architecture}

The difference in measured abstract linguistic information when learning pseudo-labels online using wav2vec 2.0 compared to second-iteration models, even given the same amount of cumulative training, suggests a possible benefit of freezing pseudo-labels during training and only updating them at fixed intervals.
The iterative refinement scheme of HuBERT and the joint learning scheme of wav2vec 2.0 may be viewed as two points on a spectrum of pseudo-label refinement: wav2vec 2.0 iterates on pseudo-labels quickly, updating them at every parameter update, whereas HuBERT iterates on pseudo-labels slowly, only updating them after an entire pass of pretraining comprising hundreds of thousands of parameter updates.
It is possible that the task of learning good pseudo-labels at the same time as good abstractions for categorization using those labels hinders wav2vec 2.0's learning because it presents a moving target during training.
This suggests that the rate of pseudo-label updating is an important parameter in the design of self-supervised models for speech.
It is also possible that the cold start at the beginning of each training iteration of HuBERT allows the model to learn more effective abstractions because it is not burdened with knowledge from training with different pseudo-labels, which may not be optimal for the new targets.
Future work may investigate the the effects of various update rates on linguistic encoding in representations and downstream performance, as well as whether the cold start is beneficial to abstract learning.

\subsection{Implications for downstream tasks}
CCA scores with phoneme identity and word identity have been shown to significantly correlate with performance on downstream tasks, including automatic speech recognition and intent classification \cite{pasad_comparative_2023}.
Representations from the final layers of later iterations of HuBERT may benefit performance in these tasks. This is particularly true for tasks which require both high-fidelity linguistic representation and absence of speaker information, such as voice conversion \cite{polyak_speech_2021}.
Tasks focusing on speaker, such as speaker diarization, may benefit more from representations from earlier iterations or earlier layers.

\subsection{Limitations}

Our conclusion that training objective has little influence on what SSL models learn is restricted to the family of models that we tested, namely HuBERT-like models consisting of a convolutional encoder and a Transformer encoder with a masked pretraining task.
%
Our conclusion is also limited to the correlations we analyzed (with phonemes, words, and speakers). These correlations are fundamental in many contexts, and are related to downstream performance. Nevertheless, other evaluations (for example, other downstream tasks) may still reveal important differences  due to the objective function.

Due to resource constraints, we tested only the \textsc{base} variants of HuBERT and wav2vec 2.0, and not the \textsc{large}.
Past work has found that different variants of these models can display somewhat different patterns  \cite{pasad_comparative_2023, choi_self-supervised_2024}.
We leave detailed investigation of the influence of model size to future work.



\section{Summary of contributions}
We have shown via a minimal test of architectural differences between wav2vec 2.0 and HuBERT that the difference between these models with respect to encoding of abstract linguistic information (words and phonemes) is due to iterative refinement, not training objective.
As the number of training iterations increases, the level of correlation to words and phonemes increases while correlation to speaker identity falls in the later layers.
We propose that the frequency of pseudo-label updating is a critical parameter in the design of self-supervised models for speech.
We suggest that future work investigate the reason for the effectiveness of iterative refinement in improving linguistic encoding, and develop techniques for for leveraging this property towards more efficient and effective speech models.

\section{Acknowledgements}
This work was supported by the Natural Sciences and Engineering Research Council of Canada (NSERC) RGPIN-2022-04431, and the Data Sciences Institute and the Linguistics Graduate Research Award at the University of Toronto, as well as resources provided by Compute Ontario, Calcul Québec, and the Digital Research Alliance of Canada.


\bibliographystyle{IEEEtran}
\bibliography{mybib}

\end{document}